\documentclass[10pt,twocolumn,letterpaper]{article}

\usepackage{iccv}
\usepackage{times}
\usepackage{epsfig}
\usepackage{graphicx}
\usepackage{amsmath}
\usepackage{amssymb}

\usepackage{multirow}
\usepackage{graphicx}
\usepackage{stackengine}
\usepackage{amssymb}
\usepackage{pifont}

\usepackage{xcolor}
\usepackage[normalem]{ulem}

\newcommand\delete{\bgroup\markoverwith{\textcolor{red}{\rule[0.5ex]{2pt}{1.0pt}}}\ULon}
\usepackage[breaklinks=true,bookmarks=false]{hyperref}

\iccvfinalcopy % *** Uncomment this line for the final submission

\def\iccvPaperID{4434} % *** Enter the ICCV Paper ID here

\ificcvfinal\fi

\begin{document}
%%%%%%%%% TITLE
\title{Dynamic Context Correspondence Network for Semantic Alignment}

\author{Shuaiyi Huang, Qiuyue Wang, Songyang Zhang, Shipeng Yan, Xuming He\\
ShanghaiTech University\\
{\tt\small $\{$huangshy1, wangqy2, zhangsy1, yanshp, hexm$\}$@shanghaitech.edu.cn}
}

\maketitle
\ificcvfinal\thispagestyle{empty}\fi

%%%%%%%%% ABSTRACT
\begin{abstract}
		
Establishing semantic correspondence is a core problem in computer vision and remains challenging due to large intra-class variations and lack of annotated data. In this paper, we aim to incorporate global semantic context in a flexible manner to overcome the limitations of prior work that relies on local semantic representations. To this end, we first propose a context-aware semantic representation that incorporates spatial layout for robust matching against local ambiguities. We then develop a novel dynamic fusion strategy based on attention mechanism to weave the advantages of both local and context features by integrating semantic cues from multiple scales. We instantiate our strategy by designing an end-to-end learnable deep network, named as Dynamic Context Correspondence Network (DCCNet). To train the network, we adopt a multi-auxiliary task loss to improve the efficiency of our weakly-supervised learning procedure.  
Our approach achieves superior or competitive performance over previous methods on several challenging datasets, including PF-Pascal, PF-Willow, and TSS, demonstrating its effectiveness and generality.

\end{abstract}

%%%%%%%%% BODY TEXT
\vspace{-3mm}

\section{Introduction}
%sem: co-seg, img editing,co-segmentaiton, robotics
%conv: image stitching
%, and enjoys wide applications such as image stitching\cite{szeliski2007image}, 3D reconstruction\cite{mustafa2017semantically} and scene recognition\cite{liu2011sift,liao2017visual}. 

% Introduce the background of the semantic correspondence task
Estimating dense correspondence across related images is a fundamental task in computer vision~\cite{scharstein2002taxonomy,hirschmuller2007stereo,horn1981determining}. While early works have focused on correspondence between images depicting the same object or scene, semantic alignment aims to find dense correspondence between different objects belonging to the same category~\cite{liu2011sift}. Such semantic correspondence has attracted much attention recently~\cite{han2017scnet, Rocco2018,kim2018recurrent} due to its potential use in a broad range of real-world applications such as image editing~\cite{dale2009image}, co-segmentation~\cite{taniai2016joint}, 3D reconstruction and scene recognition~\cite{agarwal2011building,nikandrova2015category}.  
However, this task remains extremely challenging because of large intra-class variations, viewpoint changes, background clutters and lack of data with dense annotation~\cite{Rocco2017, Rocco2018}.

% Introduce the current methods and issues for this task
There has been tremendous progress in semantic correspondence recently, thanks to learned feature representations based on convolutional neural networks (CNNs) and the adoption of weak supervision strategy in network training~\cite{Rocco18b,Rocco2018,Rocco2017,kim2018recurrent,kim2017dctm,kim2017fcss}. 
Most existing approaches learn a convolutional feature embedding so that similar image patches are mapped close to each other in the feature space, and use nearest neighbor search or geometric models for correspondence estimation~\cite{Rocco2017,Rocco2018,kim2018recurrent,kim2017dctm}. In order to achieve localization precision and robustness against deformations, such feature representations typically capture local image patterns which are insufficient to encode global semantic cues. Consequently, they are particularly sensitive to large intra-class variations and the presence of repetitive patterns. 
While recent efforts~\cite{kim2017fcss,Rocco18b} introduce local neighborhood cues to improve the matching quality, their effectiveness is limited by the local operations and short-range context.    

\begin{figure}[t]
	\centering
	\includegraphics[width=\linewidth]{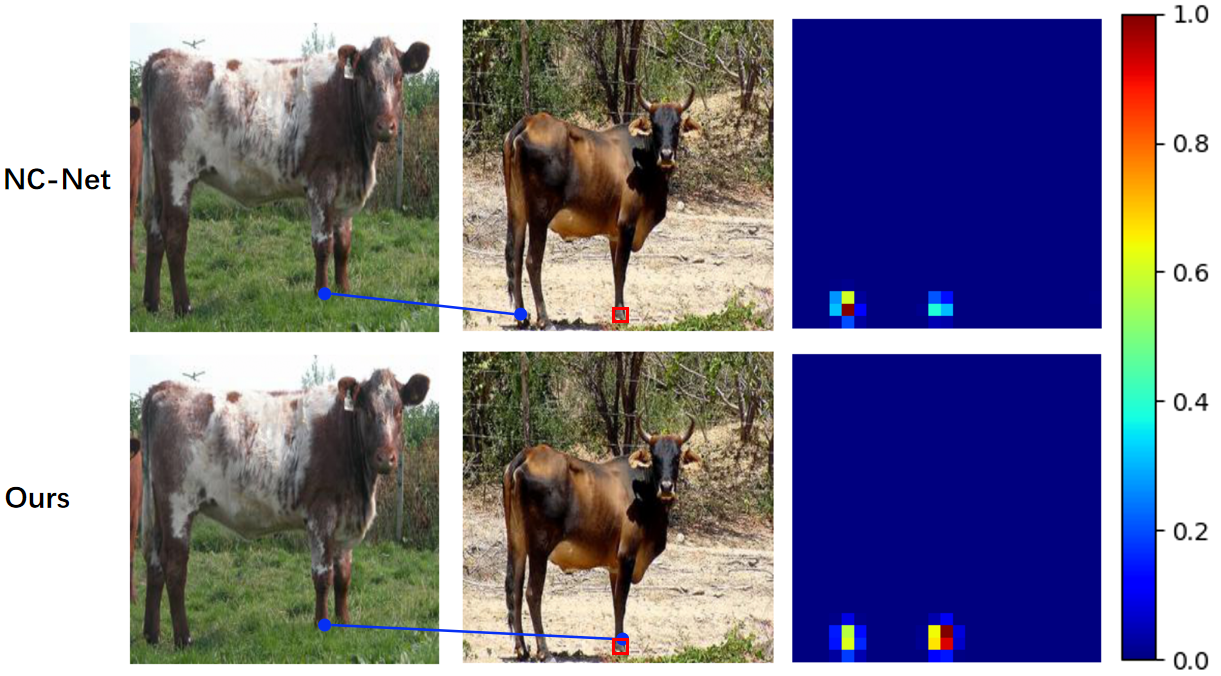}
	\caption{Given a point in the source image (blue dots in column 1), our goal is to match the corresponding point (red squares in column 2) in the target image. The values of correlation score maps (column 3) indicate the likelihood of the corresponding point locating at each location in the target image. Our model (row 2) predicts correspondence with higher precision than the baseline model \cite{Rocco18b} (row 1), demonstrating its robustness despite the repetitive patterns (blue dots in column 2).} \vspace{-3mm}
	\label{fig:vis_add}
\end{figure}

%Response heatmap between source point feature representation and target image feature map(column3), 
%Visualization of our proposed model on PF-PASCAL(the second row), demonstrating its robust matching ability despite the repetitive patterns.
 
In this work, we aim to address the aforementioned limitations by incorporating global context information and a fusion mechanism that weaves the advantages of both local and spatial features for accurate semantic matching, as shown in Fig. \ref{fig:vis_add}. To this end, we first introduce a context-aware semantic representation that integrates appearance features with a self-similarity pattern descriptor, which enables us to capture global semantic context with spatial layout cues. In addition, we propose a pixel-wise attention mechanism that dynamically combines correlation maps derived from local features and context-aware semantic features. The key idea of our approach is to reduce matching ambiguities and to improve localization accuracy simultaneously by the dynamic blending of information from multiple spatial scales.

% Introduce our methods 
Concretely, we develop a novel Dynamic Context Correspondence Network (DCCNet), which consists of three main modules: a spatial context network, a correlation network and an attention fusion network. Given an input image pair, we first compute their convolutional (conv) features using a backbone CNN (e.g., ResNet~\cite{he2016deep}). The conv features are fed into our first module, the spatial context network, which computes the context-aware semantic features that are robust against repetitive patterns and ambiguous matching. Our second module, the correlation network, has two shared branches that generates two correlation score maps for the context-aware semantic and the original conv features respectively. The third module, attention fusion net, predicts a pixel-wise weight mask to fuse two correlation score maps for final correspondence prediction. Our network is fully differentiable and is trained with a weakly-supervised strategy in an end-to-end manner. To improve the training efficiency, we propose a new hybrid loss with multiple auxiliary tasks.   

%Our approach consists of three parts and the framework is illustrated in figurexxx. First, we propose a spatial-context aware feature representation for capturing self similarity which is robust to repetitive pattern. Then, a soft attention model is designed to adaptively fuse matching predictions from different scales of feature, i.e. traditional local image appearance representation and proposed context spatial representation. Furthermore, the networks are learned via a weakly-supervised manner through a proposed dynamic loss defined between different scales to enhance regularization. Spatial context and  local feature representation 

% Validate the methods and summarize the contributions
We evaluate our method by extensive experiments on three public benchmarks, including PF-Willow~\cite{ham2016proposal}, PF-PASCAL~\cite{ham2018proposal} and TSS datasets~\cite{taniai2016joint}. The experimental results demonstrate the strong performance of our model, which outperforms the prior state-of-the-art approaches in most cases. We also conduct a detailed ablation study to illustrate the benefits of our proposed modules. 

The main contributions of this work can be summarized as follows:
\begin{itemize}\vspace{-2mm}
	\item We propose a context-aware semantic representation to generate robust matching against repetitive patterns and local ambiguities in the semantic correspondence problem.\vspace{-2mm}
	\item We develop a novel dynamic fusion strategy based on an attention mechanism to integrate multiple levels of feature representation. To the best of our knowledge, we are the first to adaptively combine context spatial information with local appearance in the semantic correspondence task.\vspace{-2mm}
	\item We design a multi-auxiliary task loss to regularize the training process for weakly-supervised semantic correspondence task and achieve superior or competitive performance on public benchmarks.
%	\item We achieve state of the art performance on various challenging datasets and demonstrate effectiveness of our modules through extensive ablation studies.
\end{itemize}

\vspace{-3mm}

\section{Related Work}

\begin{figure*}[ht]
	\centering
	\includegraphics[width=0.9\linewidth]{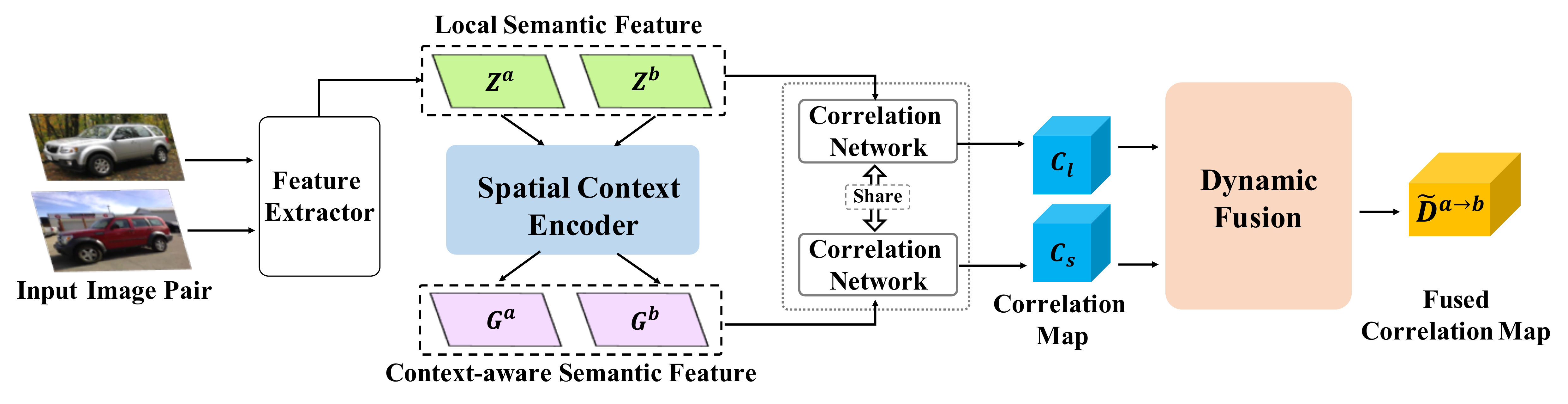}
	\caption{\textbf{Overview of DCCNet.} Our proposed DCCNet consists of three main modules: a spatial context encoder, a correlaton network and a dynamic fusion network, which are used to produce a fused correlation map.} \vspace{-3mm}
	\label{fig:overview}
\end{figure*}
%\newpage

\paragraph{Semantic Correspondence}
Traditional methods of semantic matching mostly utilize hand-crafted features to find similar image patches with additional spatial smoothness constraints in their alignment models~\cite{liu2011sift,tola2010daisy,taniai2016joint}. SIFT Flow~\cite{liu2011sift} extends classical optical flow to establish correspondences across similar scenes using dense SIFT descriptors. %Yang~\etal propose an efficient descriptor, DAISY~\cite{tola2010daisy}, to extract local features for matching. 
Taniai~\etal~\cite{taniai2016joint} adopt HOG descriptors to jointly perform co-segmentation and dense alignment. Due to lack of semantics in feature representations, those approaches often suffer from inaccurate matching when facing large appearance changes from intra-class variations.

Recently, CNNs have been successfully applied to semantic matching thanks to their learned feature representations, which are more robust to appearance or shape variations. Early attempts~\cite{novotny2017anchornet,kim2017fcss} employ learnable feature descriptors with hand-drafted alignment models, while other approaches~\cite{han2017scnet,kim2017fcss} requires external modules to generate object proposals for feature extraction, all of which are hence not end-to-end trainable. 
More recent work tends to use fully trainable network to learn the feature and alignment jointly. Rocco \etal~\cite{Rocco2017} proposes a network architecture for geometric matching using a self-supervised strategy from synthetic images, and further improves it with weakly-supervised learning in~\cite{Rocco2018}. The follow-up work extends this strategy in several directions by improving the global transformation model~\cite{hongsuck2018attentive}, developing cycle-consistency loss~\cite{chendeep}, estimating locally-varying geometric fields~\cite{kim2018recurrent, jeon2018parn}, or exploiting neighborhood consensus to produce consistent flow~\cite{Rocco18b}. 
However, most CNN-based approaches rely on dense matching of conv features, which are incapable of encoding global context~\cite{luo2016understanding,chen2017rethinking}. 
  
%\delete{WeakAlign \cite{Rocco2018} further improves this by introducing a soft inlier loss and enables weakly- supervised learning. Inspired by this, RTN \cite{kim2018recurrent} iteratively infers locally-varing geometric fields. PARN \cite{jeon2018parn} estimates locally-varing transformation fields in a coarse-to-fine manner. In this work we use NC-Net \cite{Rocco18b} as our reference implementation. This network achieves state-of-the-art performance on the PF-PASCAL \cite{ham2018proposal} dataset and impressing results on PF-WILLOW \cite{ham2016proposal} and TSS \cite{taniai2016joint} datasets.} 
%\add{A series of fully trainable network emerge since then. Among them are networks estimating a global transformation~\cite{Rocco2018, hongsuck2018attentive, chendeep}, networks estimating locally-varying geometric fields~\cite{kim2018recurrent, jeon2018parn}, and networks that do not estimate transformation directly~\cite{Rocco18b}.}
%\add{However, most above approaches are based on dense matching of local image features, which are not capable of encoding global semantic cues. While local neighborhood cues improve matching quality in more recent works~\cite{kim2017fcss, Rocco18b}, they are limited by the local operations and short-range context.}

\vspace{-3mm}
\paragraph{Spatial Context in Correspondence}

%%%%%%%%%%%%%%%%%%%%%%%%%%%%%%%% Instance-level, submission version used this paragraph
%The traditional approach~\cite{bay2006surf,hacohen2011non,harris1988combined,lowe2004distinctive} for finding the correspondences involves identifying interest points and computing local descriptors around these points with hand-craft descriptors. Though this kind of approaches perform relatively well for instance-level matching, the feature detectors and descriptors mainly rely on the local representation, and lack generalization ability for category-level matching.
%%%%%%%%%%%%%%%%%%%%%%%%%%%%%%%%%%%
%Spatial Context
Spatial context has been explored for semantic matching in the literature before deep learning era. Particularly, Irani~\etal propose the Local Self Similarity (LSS) descriptor~\cite{shechtman2007matching} to capture self-similarity structure, which has been extended to deep learning based correspondence estimation~\cite{kim2015dasc,kim2016deep}. 
%\delete{In semantic correspondence, FCSS \cite{kim2017fcss} reformulates LSS as a CNN module, and it learns a sampling pattern within a local window, which lacks of spatial context information for precise localization. Different from FCSS \cite{kim2017fcss}, our method exploits the spatial context with point-wise correlation descriptors based on self-similarity, in which we are able to leverage the spatial context at different scales, and generate robust feature representations against the class variations.}
More recent work, FCSS~\cite{kim2017fcss} and its extension~\cite{kim2017dctm}, reformulate LSS as a CNN module, computing local self-similarity with learned sparse sampling pattern in object proposals. 
In contrast, our method exploits a larger spatial context and computes a dense self-similarity descriptor, which is more robust against repetitive patterns and encodes richer context. %Generating different levels of spatial context descriptor can be easily achieved in our operator by setting different kernel size. 
We also combine this descriptor with local conv features, further improving the discriminative capability of our feature and stabilizing training.

\vspace{-3mm}
\paragraph{Dynamic Fusion}
Attention mechanism has been widely used in computer vision tasks to focus on relevant information. For instance, attention-based dynamic fusion is adopted for confidence measure in stereo matching~\cite{kim2019laf}. In semantic segmentation, Chen~\etal~\cite{chen2016attention} propose an attention mechanism that learns to fuse multi-scale features at each pixel location. In semantic correspondence, 
%\delete{recent methods design attention modules for determining foreground or background~\cite{chendeep}, features reliable or not~\cite{hongsuck2018attentive}} 
recent methods design attention modules for suppressing background regions in images~\cite{chendeep, hongsuck2018attentive}. By contrast, our work addresses the challenge of integrating local and context cues in semantic matching, for which, to the best of our knowledge, dynamic fusion has not been explored before. 

\vspace{-3mm}

\newcommand\oast{\stackMath\mathbin{\stackinset{c}{0ex}{c}{0ex}{\ast}{\bigcirc}}}
%\vspace{-5mm}

\section{Method}

% Overview
%hsy
We now describe our method for estimating a robust and accurate semantic correspondence between two images. Our goal is to seek a flexible feature representation that enables us to capture global semantic contexts as well as informative local features. To this end, we introduce a learnable context-aware semantic representation that augments each local convolutional feature with a global context descriptor. Such a context-aware feature is integrated into the correlation computation by a dynamic fusion mechanism, which combines correlation scores from the context-aware feature and the local conv feature in a selective manner to generate high-quality matching predictions.  

Below we start a brief introduction to the semantic correspondence task and an overview of our framework in Sec.~\ref{subsec:problem}. We then present our proposed context-aware semantic feature and its encoder network in Sec.~\ref{subsec:spatial}, followed by a dynamic fusion module in Sec.~\ref{subsec:dynamic}. Finally, we describe our multi-auxiliary task loss in Sec.~\ref{subsec:aux_loss}. 

\subsection{Problem Setting and Overview}\label{subsec:problem}

\begin{figure*}[ht]
	\centering
	\includegraphics[width=0.75\linewidth]{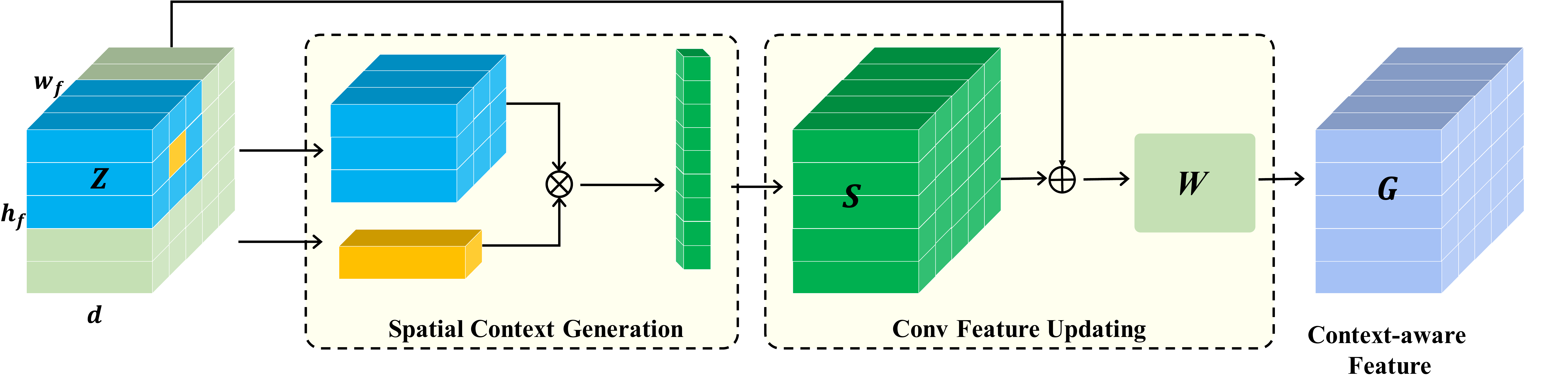}
	\vspace{-1mm}
	\caption{\textbf{Overview of Spatial Context Encoder.} Spatial context encoder generates context-aware features from local conv features. }\vspace{-3mm}
	\label{fig:SCE}
\end{figure*}

% Framework of current method
Given an input image pair $(\mathbf{I}^a,\mathbf{I}^b)$, the goal of semantic alignment is to estimate a dense 
correspondence between pixels in two images. A common strategy is to infer the correspondence from a correlation map $\mathbf{C}^I$, which describes the matching similarities between any two locations from different images. Formally, let $\mathbf{I}^a\in\mathbb{R}^{3\times h^a \times w^a}, \mathbf{I}^b\in\mathbb{R}^{3\times h^b \times w^b}$, where $h^a, h^b$ and $w^a, w^b$ are the height and width of those two images, respectively. The correlation map is denoted as $\mathbf{C}^I\in\mathbb{R}^{h^a \times w^a \times h^b \times w^b}$ and $\mathbf{C}^I_{(i,j,m,n)}= f(I_{(i,j)}^a,I_{(m,n)}^b)$ where $f$ is a similarity function. 
To achieve point-to-point spatial correspondence between two images, we can perform a hard assignment in either of two possible directions, from $\mathbf{I}^a$ to $\mathbf{I}^b$, or vice versa (cf.~\cite{Rocco18b}). Specifically, we have the following mapping from $a$ to $b$:
\begin{align}
&\mathbf{I}_{(i,j)}^a \text{ correspond to a given } \mathbf{I}_{(m,n)}^b \nonumber \\\Leftrightarrow\quad& (i,j)=\mathop{\arg\max}_{1\leq i'\leq h^a,1\leq j'\leq w^a}\mathbf{C}^I_{(i',j',m,n)}\label{eq:hard}
\end{align}
By doing so, we convert the semantic correspondence problem to a correlation map prediction task, in which our goal is to find a functional mapping from the image pair to an optimal correlation map that generates the accurate pixel-wise correspondences. 

A typical deep learning based approach aims to build a high-quality correlation map based on learned feature representation. Formally, we first compute the conv features of the images $\mathbf{I}^a,\mathbf{I}^b$ by an embedding network, which is pre-trained on a large dataset (e.g., ImageNet). Denoting the embedding network as $\mathcal{F}$, we generate the image feature maps as follows,
\begin{align}
	\mathbf{Z}^a = \mathcal{F}(\mathbf{I}^a),\quad\mathbf{Z}^b = \mathcal{F}(\mathbf{I}^b),
\end{align}  
where $\mathbf{Z}^a\in\mathbb{R}^{d\times h^a_f\times w^a_f}$ and $\mathbf{Z}^b\in\mathbb{R}^{d\times h^b_f\times w^b_f}$ are the normalized conv feature representations of the input image pair $(\mathbf{I}^a,\mathbf{I}^b)$, $d$ is the number of feature channel.

Given the conv features, we then build a correlation network that learns a mapping from the feature pair to their correlation map $\mathbf{C} \in \mathbb{R}^{h_f^a\times w^a_f\times h^b_f\times w^b_f}$. Formally,  
\begin{align}
\mathbf{C}  = \mathcal{G}(\mathbf{Z}^a,\mathbf{Z}^b;\Theta^{ab})	
\end{align}
where $\mathcal{G}$ is the mapping function implemented by the deep network and $\Theta_{ab}$ is its parameters. Given a feature-wise correspondence, we can derive the pixel-wise correspondences in Eq.~\eqref{eq:hard} by interpolation on the image plane. 

While such deep correspondence networks (e.g.~\cite{Rocco18b}) provide a powerful framework to learn a flexible representation for matching, in practice they are sensitive to large intra-class variations and repetitive patterns in images due to lack of global context.    
% Pipeline of our method
In this work, we propose a novel correspondence network to tackle those challenges in semantic correspondence. Our network is capable of capturing global context of each feature location and dynamically integrating context-aware semantic cues with local semantic information to reduce the matching ambiguities. Hence we refer to our network as Dynamic Context Correspondence Network (DCCNet). Our DCCNet network is composed of three main modules: 1) a \textit{spatial context} encoder, 2) a \textit{correlation} network and 3) a \textit{dynamic fusion} network. Below we will introduce the details of each module and an overview of our network is illustrated in Fig.~\ref{fig:overview}.

%It is worth noting that we adopt the neighborhood consensus module proposed by \cite{Rocco18b} in the third stage, thus we will focus on the  \textit{spatial context encoder} and \textit{dynamic context fusion stage} in the following sections.

\subsection{Spatial Context Encoder}\label{subsec:spatial}
Taking as input the conv features of the image pairs, the first component of DCCNet is a spatial context encoder that incorporates global semantic context into the conv feature. %Our goal is to generate a context-aware semantic representation for each location on the feature map. 
To achieve this, we propose a self-similarity based operator to describe the spatial context, as shown in Fig.~\ref{fig:SCE}. Specifically, the spatial context encoder consists of two modules: a) \textit{spatial context generation}, b) \textit{context-aware semantic feature generation}, which will be detailed below.

%our spatial context descriptor captures global feature patterns in a window of size $k\times k$ centered at each feature location.%, where $k$ indicates the spatial scale of the descriptor. %By varying $k$, we are able to encode different levels of spatial context.   

\vspace{-2mm}
\paragraph{Spatial Context Generation} Inspired by LSS~\cite{shechtman2007matching}, we design a novel self-similarity based descriptor on top of deep conv features to encode spatial context at each location in an image. 
Concretely, given the conv feature map $\mathbf{Z}= \{\mathbf{z}_{(i,j)}\}\in\mathbb{R}^{d\times h_f \times w_f}$ of an image $\mathbf{I}$ (omit superscript here for clarity), we first apply a zero padding of size $(k-1)/2$ ($k$ is odd) on the feature map $\mathbf{Z}$ to get the padded feature map $\widetilde{\mathbf{Z}}\in\mathbb{R}^{d\times (h_f+k-1)\times(w_f+k-1)}$. 
For location ($i,j$) in $\mathbf{Z}$, its spatial context descriptor is defined as a self-similarity vector computed between its own local feature $\mathbf{z}_i$ and the features in its neighboring region of size $k\times k$ centered at ($i,j$). Specifically, we compute the self-similarity features as follows:
\begin{align}
%	\mathbf{s} = [\mathbf{z}^p_{m,n}\mathbf{z}_{i,j}^\top,\mathbf{z}^p_{m,n}\mathbf{z}_{i,j}^\top,\mathbf{z}^p_{m,n}\mathbf{z}_{i,j}^\top]
	&\mathbf{s}_{(i,j)} = %\mathbf{z}^\intercal_{i,j}[\tilde{\mathbf{z}}_{(i,j)},\cdots,\tilde{\mathbf{z}}_{(i+k,j+k)}]\\
	%&\qquad= 
	[\mathbf{z}^\intercal_{i,j}\tilde{\mathbf{z}}_{(i,j)},\cdots,\mathbf{z}^\intercal_{i,j}\tilde{\mathbf{z}}_{(i+k-1,j+k-1)}], \\
	&\mathbf{S} = \{\mathbf{s}_{(1,1)},\cdots,\mathbf{s}_{(h_f,w_f)}\},\\
	&\mathbf{s}_{(i,j)}\in\mathbb{R}^{k^2\times 1}, \quad\mathbf{S}\in\mathbb{R}^{k^2\times h_f\times w_f},
\end{align}
where $\mathbf{s}_{(i,j)}$ is the spatial context descriptor of location $(i,j)$ and $\mathbf{S}$ denotes spatial context of the image $I$. We refer the neighborhood size $k$ as the kernel size of the context descriptor. With varying kernel sizes, the descriptor is able to encode the spatial context at different scales.

It is worth noting that our spatial context descriptor differs from non-local graph networks~\cite{wang2018non} in encoding context information, as our descriptor maintains spatial structure, which is important for matching, while graph propagation typically uses aggregation operators to integrate out spatial cues. %We argue that maintaining structure information is rather important in semantic matching where large intra-class variations and appearance changes exist, as visual property like color may not share within two images but geometric layout can. 
Our representation also differs from FCSS \cite{kim2017fcss} in several aspects. First, we use a large context to compute self-similarity instead of a local window in order to achieve robustness toward repetitive patterns. %This can be easily achieved by adjusting kernel size in our operator while FCSS~\cite{kim2017fcss} learned a sampling pattern within a local window. 
Second, FCSS~\cite{kim2017fcss} relies on object proposals to remove background while we learn to select informative semantic cues. Moreover, we empirically find that the spatial context descriptor alone is insufficient for high-quality matching, and therefore combine it with local conv features, which will be described below.

%\paragraph{Conv Feature Augmentation.}
\vspace{-2mm}
\paragraph{Context-aware Semantic Feature}
The second module of our spatial context encoder computes a context-aware semantic feature for each location on the conv feature map. While the spatial context descriptor encodes second-order statistics in a neighborhood of feature location, it lacks local semantic cues represented by the original conv feature. In order to capture different aspects of semantic objects, we employ a simple fusion step to generate a context-aware semantic representation which provides us better matching quality. Concretely, we apply a non-linear transformation over the concatenation  of $\mathbf{Z}$ and $\mathbf{S}$ as below:
\begin{align}
	&\mathbf{G}_{(i,j)}= \sigma(\mathbf{W}^\intercal[\mathbf{s}^\intercal_{(i,j)},\mathbf{z}^\intercal_{(i,j)}]^\intercal)\\
	&\mathbf{G} = \{{\mathbf{g}}_{(1,1)},\cdots,{\mathbf{g}}_{(h_f,w_f)}\}\\
	&{\mathbf{g}}_{(i,j)}\in\mathbb{R}^{l},\quad {\mathbf{G}}\in\mathbb{R}^{l\times h_f\times w_f}
\end{align}
where $\sigma$ is a nonlinear function (ReLU) and the weight matrix $\mathbf{W}\in\mathbb{R}^{(d+k^2)\times l}$ transforms the features into $l$ dimensional space. We use $\mathbf{G}$ to denote the context-aware semantic features of image $\mathbf{I}$, and add superscript to represent context-aware semantic feature ${\mathbf{G}}^a$ and ${\mathbf{G}}^b$ from the image $\mathbf{I}^a$ and $\mathbf{I}^b$, respectively.

\begin{figure}
\begin{center}
%\fbox{\rule{0pt}{2in} \rule{.9\linewidth}{0pt}}
%\includegraphics[scale=0.3]{figures/loss.png}
\includegraphics[width=0.9\linewidth]{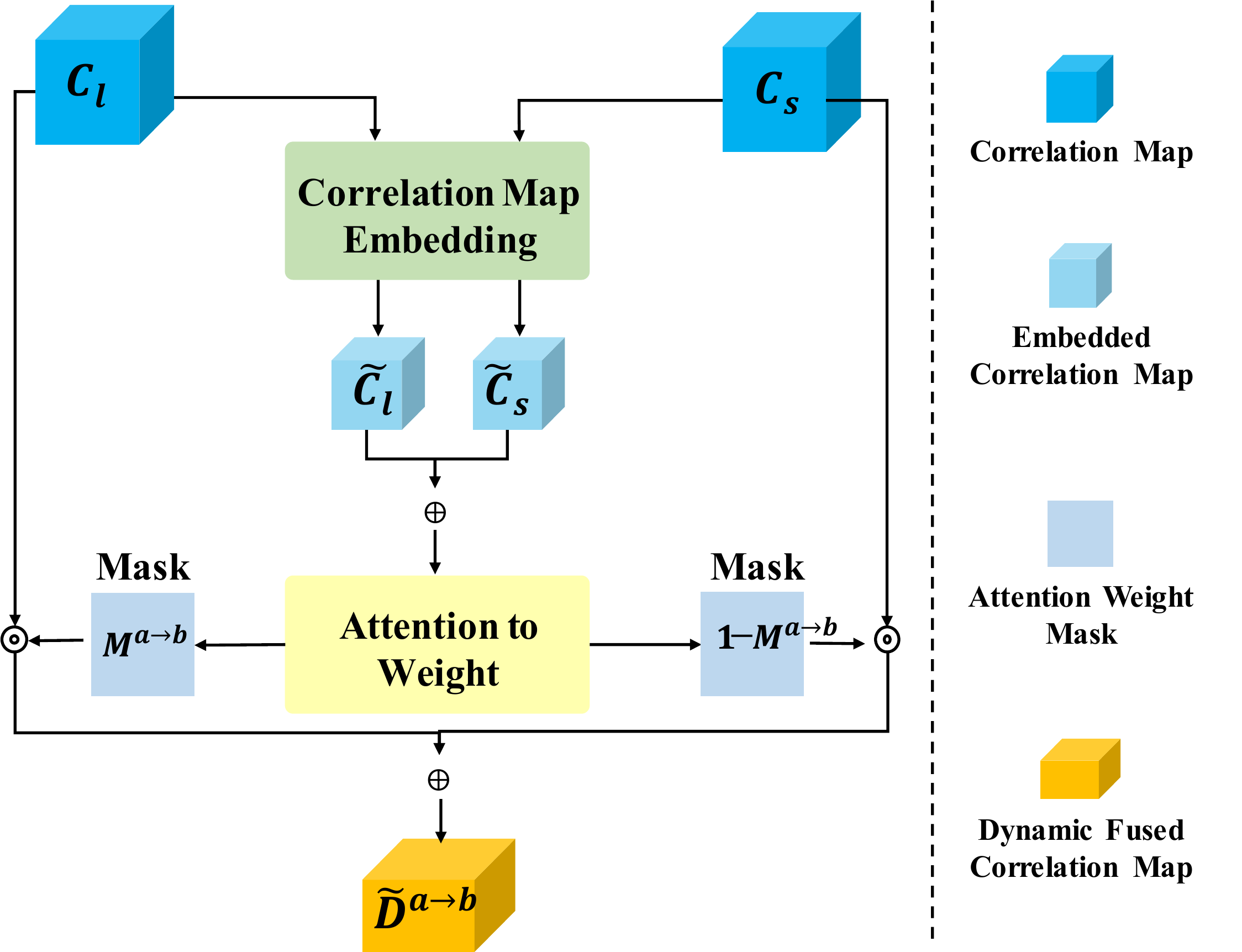}
\end{center}
\vspace{-3mm}
   \caption{\textbf{Overview of Dynamic Fusion Network.} The network employs correlation map embedding and attention-based fusion to combine context and local semantic cues.}\vspace{-2mm}
\label{fig:dynamic_fusion}
\end{figure}

\subsection{Correlation Network}\label{subsec:corrnet}

The second module of DCCNet is a correlation network that takes in feature representations of an image pair and produces a correlation map. While any correlation computation module can be used here, we adopt the neighborhood consensus module~\cite{Rocco18b} in this work for its superior performance. Specifically, for each type of feature representations of an image pair, say the context-aware semantic feature $(\mathbf{G}^a,\mathbf{G}^b)$ or the local semantic feature $(\mathbf{Z}^a,\mathbf{Z}^b)$, we feed them into the correlation network to generate their corresponding correlation map: 
\begin{align}
 &\mathbf{C}_l=\mathcal{H}(\mathbf{Z}^a \oast \mathbf{Z}^b),\quad \mathbf{C}_s=\mathcal{H}({\mathbf{G}}^a \oast {\mathbf{G}}^b)\\
 &\mathbf{C}_l, \mathbf{C}_s\in\mathbb{R}^{h^a_f\times w^a_f\times h_f^b\times w_f^b}
\end{align}
where $\mathcal{H}$ is the neighborhood consensus operator, $\oast$ is the correlation operation. We use $\mathcal{H}$ to refine the correlation maps based on local neighborhood information. In addition, mutual nearest neighbor consistency constraint~\cite{Rocco18b} is applied before and after $\mathcal{H}$, which is merged into $\mathcal{H}$ for simplicity as it does not contain learnable parameters. We refer the reader to \cite{Rocco18b} for more details.
We now have two correlation maps, $\mathbf{C}_s$ and $\mathbf{C}_l$, that describes the pixelwise correspondence using context-aware semantic cues and local semantic features, respectively.

%More specifically, the correlation maps of the context semantic feature and the local semantic feature are computed through the neighborhood consensus module\cite{Rocco18b} separately first, the output of which can be denoted as $\mathbf{C}_s$ and $\mathbf{C}_l$, respectively.
% \begin{align}
% 	&\mathbf{C}_l=\mathcal{H}(\mathbf{Z}^a,\mathbf{Z}^b),\quad \mathbf{C}_s=\mathcal{H}(\mathbf{S}^a,\mathbf{S}^b)\\
% 	&\mathbf{C}_l, \mathbf{C}_s\in\mathbb{R}^{h^a_f\times w^a_f\times h_f^b\times w_f^b}
% \end{align}

% $\mathbf{C}_l$ is computed from the original conv feature representation, which modeling the pairwise correlation of the image pairs at the local semantic level. $\mathbf{C}_s$ is predicted with the spatial context augmented feature, which is focusing on the correlation in the level of spatial context.

\subsection{Dynamic Fusion Network}\label{subsec:dynamic}

While the context-aware semantic feature allows us to encode more global visual patterns, the spatial context encoder in Sec.~\ref{subsec:spatial} adopts a spatial-invariant fusion mechanism (i.e., a global embedding) to combine local cues and spatial context, which turns out to be sub-optimal for feature locations with distracting neighboring region. An effective solution is to introduce a spatially varying fusion mechanism to balance the context and local conv features specifically for each location. To that end, we propose a dynamic fusion strategy to achieve adaptive fusion for different locations in each image pair. Our fusion utilizes scores from two correlation maps computed in Sec.~\ref{subsec:corrnet} for each location and determines which one is more trustworthy using a location-specific weight.  

Specifically, given two correlation maps, $\mathbf{C}_s$ and $\mathbf{C}_l$, we introduce the third module of DCCNet, a dynamic fusion network, to integrate two correlation scores. %\add{In Sec.~\ref{subsec:spatial}, our spatial context encoder combines local cues and spatial context cues by an embedding shared by all the locations, which does not provide a spatially varying fusion mechanism.} Our goal \add{here} is to balance these two kinds of correlation measurement for each location in an image pair to achieve more precise localization and matching. Toward this objective, we propose a dynamic fusion strategy to achieve adaptive fusion for different locations of each image pair. \add{It utilizes scores from two correlation maps respectively for each location and determines which one is more trustworthy using a location-specific weight.}
Motivated by \cite{chen2016attention}, we exploit an attention mechanism to generate a location-aware weight mask for correlation map fusion. The attention-based dynamic fusion consists of the following two modules: 1) \textit{correlation map embedding}, 2) \textit{attention-based fusion}, which will be described below.  

Our dynamic fusion strategy is associated with the matching direction. Here we describe the dynamic fusion in the direction from image $\mathbf{I}^a$ to image $\mathbf{I}^b$ for clarity, as the other direction is similar, as shown in Fig.~\ref{fig:dynamic_fusion}.

  \vspace{-3mm}
\paragraph{Correlation Map Embedding} In order to predict the attention mask, we first compute a feature representation from the correlation maps. Concretely, we apply an embedding function $\mathcal{E}$ to produce a correlation map embedding:
 \begin{align}
 	\tilde{\mathbf{C}}_s = \sigma(\mathcal{E}(\mathbf{C}_s;\theta_\mathcal{E})),\quad \tilde{\mathbf{C}}_l = \sigma(\mathcal{E}(\mathbf{C}_l;\theta_\mathcal{E}))
 \end{align}
where $\mathcal{E}$ is implemented by 4D convolutional neural network, and $\theta_{\mathcal{E}}$ is the learnable parameter of $\mathcal{E}$. $\tilde{\mathbf{C}}_s, \tilde{\mathbf{C}}_l $ are at the same dimension with ${\mathbf{C}}_s, {\mathbf{C}}_l$, in $\mathbb{R}^{h_f^a\times w_f^a\times h_f^b\times w_f^b}$. By this module, we extract those 4D correlation features $\tilde{\mathbf{C}}_l$, $\tilde{\mathbf{C}}_s$, before reshaping them in the next attention module that produces the weight mask and fusion result.
 
 \vspace{-3mm}

\paragraph{Attention-based Fusion}
% We first reshape $\tilde{\mathbf{C}}_s,\tilde{\mathbf{C}}_l$ into a matrix form $\mathbf{D}_l \in \mathbb{R}^{N_a\times N_b}$ and $\mathbf{D}_s \in \mathbb{R}^{N_a\times N_b}$, where $N_a=h_f^a\times w_f^a$ and $N_b = h_f^b\times w_f^b$. 

To compute the attention weight mask, we first reshape $\tilde{\mathbf{C}}_s,\tilde{\mathbf{C}}_l$ into a tensor form $\mathbf{D}_l \in \mathbb{R}^{N_b\times h_f^a \times w_f^a}$ and $\mathbf{D}_s \in \mathbb{R}^{N_b\times h_f^a \times w_f^a}$, where $N_b = h_f^b\times w_f^b$.  
We then compute a fusion weight map for each image pair, which indicates whether the local conv feature is more informative than the context-aware semantic feature for each location. For the direction of image $\mathbf{I}^a$ to $\mathbf{I}^b$, we stack the reshaped correlation maps $\mathbf{D}_l$ and $\mathbf{D}_s$ along the first axis followed by an attention network to predict the fusion weights:
\begin{align}
	&\mathbf{D}^{a\rightarrow b}=\mathbf{D}_l\oplus \mathbf{D}_s,\quad\mathbf{D}\in\mathbb{R}^{(2N_b)\times h_f^a \times w_f^a}\\
	&\mathbf{M}^{a\rightarrow b} = \mathcal{M}( \mathbf{D}^{a\rightarrow b}),\quad \mathbf{M}^{a\rightarrow b}\in\mathbb{R}^{1\times h_f^a \times w_f^a}
%	&\mathbf{M}_b = g([\mathbf{D}^\intercal_l,\mathbf{D}^\intercal_s]),\quad \mathbf{M}_b \in \mathbb{R}^{N_b\times2}
\end{align}
where $\oplus$ is concatenation operator along the first dimension, and $\mathbf{M}^{a\rightarrow b}$ is the attention weight mask for $\tilde{\mathbf{C}}_l$. The attention network $\mathcal{M}(\cdot)$ is implemented by a fully convolution layer followed by a softmax operator to normalize the attention weights. Given the attention mask, we fuse the correlation maps in an adaptive way as follows,
\begin{align}
	&\tilde{\mathbf{D}}^{a\rightarrow b} = \mathbf{D}_l\circ\mathbf{M}^{a\rightarrow b}+ \mathbf{D}_s\circ(1-\mathbf{M}^{a\rightarrow b})\\
	&\tilde{\mathbf{D}}^{a\rightarrow b}\in\mathbb{R}^{N_b\times h_f^a \times w_f^a}
\end{align}
where $\circ$ is the element-wise multiplication with broadcasting for producing the weighted correlation maps.
The output correlation $\tilde{\mathbf{C}}^{a\rightarrow b}$ is generated by reshaping $\tilde{\mathbf{D}}^{a\rightarrow b}$ into the 4D form $\mathbb{R}^{h_f^a\times w_f^a\times h_f^b\times w_f^b}$. Similarly, the adaptively fused correlation $\tilde{\mathbf{C}}^{b\rightarrow a} \in \mathbb{R}^{h_f^a\times w_f^a\times h_f^b\times w_f^b}$ from the other direction can also be computed by this module. Finally, those two refined correlation map  $\tilde{\mathbf{C}}^{a\rightarrow b}$ and $\tilde{\mathbf{C}}^{b\rightarrow a}$ are used to find semantic correspondence (cf.~\cite{Rocco18b}).

%denotes the fused map as $\mathbf{C}^a_f$ and $\mathbf{C}^b_f$ for the image pairs, which can be generated by an element-wise multiplication and reshape operation, which we will describe[TODO]. 

%\subsection{Learning with Multi-auxiliary Loss}\label{subsec:aux_loss}

%\begin{figure}[ht]
%	\centering
%	\includegraphics[scale=0.3]{figures/attentionM.png}
%	\caption{Multi-auxiliary Loss}
%	\label{fig:aux_loss}
%\end{figure}

% Discussion
%\add{Our spatial context encoder combines local cues and spatial context cues. This combination uses an embedding shared by all the locations and thus does not provide a spatially varying fusion mechanism. In contrast, our dynamic fusion smartly combines the strengths of local features and context-aware semantic features: it utilizes correlation scores from them respectively for each location and determines which one is more trustworthy using a location-specific weight. We give more detailed analysis in our ablation study.}

% Note that the attention is estimated based on the correlation score maps $(\mathbf{C}_l, \mathbf{C}_s)$ and thus cannot be applied to computing $\mathbf{C}_s$ itself.

\newcommand{\cmark}{\ding{51}}%
\newcommand{\xmark}{\ding{55}}%
\begin{table*}[t]
	\centering
	\resizebox{1.0\textwidth}{!}{
		\begin{tabular}{ccccccccccccccccccccc|c}
			\hline
			%   & \multicolumn{3}{c}{Default} & \multicolumn{3}{c}{Know  Object}\\
			Methods  &aero&bike&bird&boat&bottle&bus&car&cat&chair&cow&d.table&dog&horse&moto&person&plant&sheep&sofa&train&tv&all  \\  
			\hline
			\hline
			HOG+PF-LOM\cite{ham2016proposal} &73.3&74.4&54.4 &50.9&49.6&73.8 &72.9&63.6&46.1&79.8 &42.5&48.0&68.3 &66.3&42.1&62.1 &65.2&57.1&64.4 &58.0&62.5     \\
			UCN\cite{choy2016universal}  &64.8&58.7&42.8&59.6&47.0&42.2&61.0&45.6&49.9&52.0&48.5&49.5&53.2&72.7&53.0&41.4&\textbf{83.3}&49.0&73.0&66.0&55.6\\
			VGG-16+SCNet-A\cite{khan2017} &67.6&72.9&69.3&59.7&74.5&72.7&73.2&59.5&51.4&78.2&39.4&50.1&67.0&62.1&69.3&68.5&78.2&63.3&57.7&59.8&66.3\\
			VGG-16+SCNet-AG\cite{khan2017}  &83.9&81.4&70.6&62.5&60.6&81.3&81.2&59.5&53.1&81.2&\textbf{62.0}&58.7&65.5&73.3&51.2&58.3&60.0&69.3&61.5&80.0&69.7\\
			VGG-16+SCNet-AG+\cite{khan2017}            &85.5&84.4&66.3&70.8&57.4&82.7&82.3&71.6&54.3&\textbf{95.8}&55.2&59.5&68.6&75.0&56.3&60.4&60.0&\textbf{73.7}&66.5&76.7&72.2\\
			VGG-16+CNNGeo\cite{Rocco2017}   &79.5&80.9&69.9&61.1&57.8&77.1&84.4&55.5&48.1&83.3&37.0&54.1&58.2&70.7&51.4&41.4&60.0&44.3&55.3&30.0&62.6\\
			ResNet-101+CNNGeo(S)\cite{Rocco2017}   &82.4&80.9&85.9&47.2&57.8&83.1&92.8&86.9&43.8&91.7&28.1&76.4&70.2&76.6&68.9&65.7&80.0&50.1&46.3&60.6&71.9\\
			ResNet-101+CNNGeo(W)\cite{Rocco2018}  &83.7&88.0&83.4&58.3&68.8&\textbf{90.3}&92.3&83.7&47.4&91.7&28.1&76.3&\textbf{77.0}&76.0&71.4&\textbf{76.2}&80.0&59.5&62.3&63.9&75.8\\
			RTN \cite{kim2018recurrent}  &-&-&-&-&-&-&-&-&-&-&-&-&-&-&-&-&-&-&-&-&75.9\\
			%    Accv\cite{chendeep}   &85.6&89.6&82.1&83.3&85.9&92.5&93.9&80.2&52.2&85.4&55.2&75.2&64.0&77.9&67.2&73.8&100.0&65.3&69.3&61.1&78.0\\
			NC-Net\cite{Rocco18b}  &86.8&86.7&\textbf{86.7}&55.6&82.8&88.6&93.8&87.1&54.3&87.5&43.2&82.0&64.1&79.2&71.1&71.0&60.0&54.2&75.0&\textbf{82.8}&78.9\\
			\hline
			Our Method&\textbf{87.3}&\textbf{88.6}&82.0&\textbf{66.7}&\textbf{84.4}&89.6&\textbf{94.0}&\textbf{90.5}&\textbf{64.4}&91.7&51.6&\textbf{84.2}&74.3&\textbf{83.5}&\textbf{72.5}&72.9&60.0&68.3&\textbf{81.8}&81.1&\textbf{82.3}\\
			\hline
		\end{tabular}}
		\caption{\textbf{Performance on the PF-Pascal dataset~\cite{ham2018proposal}}. Per-class and overall PCK are shown in the table and the best results are in bold.}
		\label{EPascal}
	\end{table*}
	
\subsection{Learning with Multi-auxiliary Task Loss}\label{subsec:aux_loss}

We learn the model parameters of our DDCNet in a weakly-supervised manner from a set of matched images. Given two images $\mathbf{I}^a$ and $\mathbf{I}^b$, the outputs of our model are $\tilde{\mathbf{C}}^{a\rightarrow b}$ and $\tilde{\mathbf{C}}^{b\rightarrow a}$. 
%\delete{We first adopt the weakly-supervised loss as in \cite{Rocco18b}, which has a functional form $\mathcal{L}_{NC}(\tilde{\mathbf{C}}^{b\rightarrow a}, \tilde{\mathbf{C}}^{a\rightarrow b}, y)$ where $y=+1$ for positive pairs, and $y=-1$ for negative pairs. We denote this loss term as $\mathcal{L}_{fuse}(\mathbf{I}^a, \mathbf{I}^b)$. } 
We first adopt the weakly-supervised training loss proposed in NC-Net \cite{Rocco18b}, which has a functional form :
	%%%%%%
	\begin{align}
%	\mathcal{L}\left(I^{a}, I^{b}\right)= \mathcal{L}_{NC}(\tilde{\mathbf{C}}^{b\rightarrow a}, \tilde{\mathbf{C}}^{a\rightarrow b}, y) =-y\left(\overline{s}^{a}+\overline{s}^{b}\right)
	\mathcal{L}(\tilde{\mathbf{C}}^{b\rightarrow a}, \tilde{\mathbf{C}}^{a\rightarrow b}, y) =-y\left(\overline{s}^{a}+\overline{s}^{b}\right)
	\end{align}
	where $y$ denotes the groundtruth label of the image pair \(\left(I^{a}, I^{b}\right)\) with $y=+1$ for positive matching, and $y=-1$ for negative. \(\overline{s}^{a}\) and \(\overline{s}^{b}\) are the mean matching scores over all hard assigned matches of a given image pair \(\left(I^{a}, I^{b}\right)\) in both matching directions. To minimize this loss, the model should maximizes the scores of positive and minimizes the scores of negative matching pairs, respectively. We denote this loss term as $\mathcal{L}_{fuse}(\mathbf{I}^a, \mathbf{I}^b)$.

To learn an effective dynamic fusion strategy, we further introduce additional supervision from two auxiliary tasks. Specifically, we also use the correlation map $\mathbf{C}_l$ of local semantic feature and the correlation map $\mathbf{C}_s$ of context-aware semantic feature to generate the matching results, and denote their correspondence losses as $\mathcal{L}_{local}$ and $\mathcal{L}_{context}$, respectively. Here we compute the auxiliary task losses $\mathcal{L}_{local}$ and $\mathcal{L}_{context}$ following the same procedure as in $\mathcal{L}_{fuse}$. The overall training loss is then defined as,
\begin{align}
	\mathcal{L}(\mathbf{I}^a, \mathbf{I}^b) = \mathcal{L}_{fuse} + \lambda \mathcal{L}_{local} + \gamma\mathcal{L}_{context}
\end{align}
where $\lambda$ and $\gamma$ are the hyper-parameter to balance the main and auxiliary task losses.

\vspace{-1mm}

\section{Experiments}
%, including NC-Net\cite{Rocco2018}, WeakAlign\cite{Rocco2018}, RTN\cite{kim2018recurrent}, FDCC\cite{chendeep}(TODO don't know its name), CNNGeo\cite{Rocco2017}, SCNet\cite{khan2017} and early work. 
We evaluate our DCCNet on the weakly-supervised semantic correspondence task by conducting a series of experiments on three public datasets, including PF-PASCAL \cite{ham2018proposal}, PF-WILLOW \cite{ham2016proposal} and TSS \cite{taniai2016joint}. In this section, we introduce our experiment settings and report evaluation results in detail.
We first describe the implementation details in Sec.\ref{sec:implementation}, followed by the quantitative results of the three datasets in Sec.\ref{sec:pf-pascal-benchmark}, Sec.\ref{sec:pf-willow-benchmark} and Sec.\ref{sec:tss-benchmark}, respectively. Finally, ablation study and comprehensive analysis are provided in Sec.\ref{sec:ablation-study}. 

\begin{figure}[t!]
	\centering
	\includegraphics[width=\linewidth]{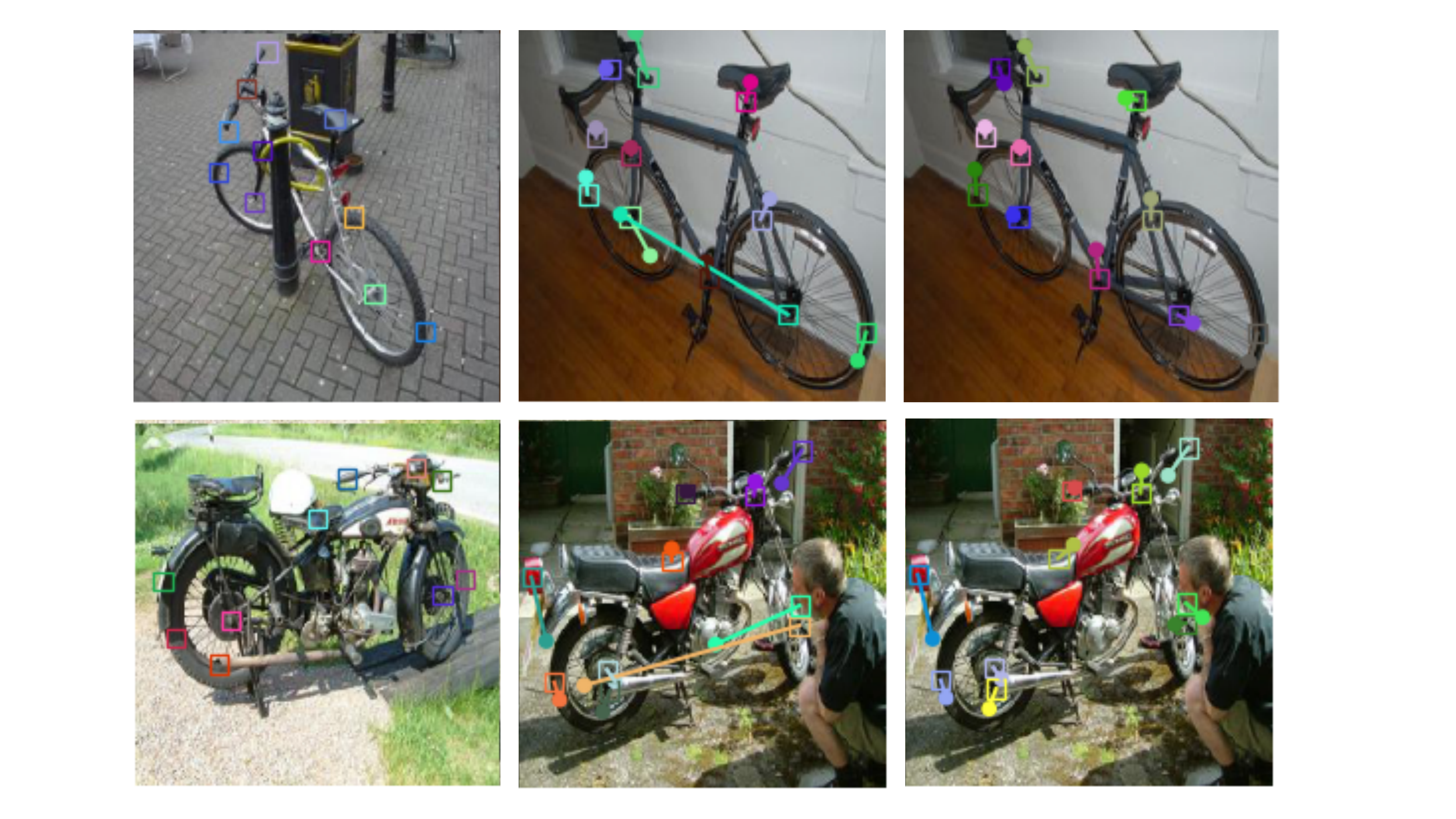}
	\caption{\textbf{Qualitative comparisons on the PF-PASCAL benchmark~\cite{ham2018proposal}.} The leftmost column shows source images. The second and third columns show predictions from Nc-Net \cite{Rocco18b} and our proposed DCCNet respectively. We show the ground truth keypoints in squares and the predicted keypoints in dots, with their distance in target images depicting the matching error. It is clear that our model is robust to repetitive patterns.} \vspace{-3mm}
	%(Best viewed in color)
	\label{fig:vis_pfpascal}
\end{figure}

\subsection{Implementation details}
\label{sec:implementation}
We implement our DCCNet with the PyTorch framework~\cite{paszke2017automatic}. For the feature extractor, we use the ResNet-101~\cite{he2016deep} pre-trained on ImageNet with the parameters fixed and truncated at the conv4\_23 layer. The spatial context encoder adopts a kernel size $k=25$ and the output dimension $l$ of the context-aware semantic features is set to 1024, which are determined by validation. For the correlation network, we follow~\cite{Rocco18b} and stack three 4D convolutional layers with the kernel size at 5$\times$5$\times$5$\times$5 and set the channel number of the intermediate layer to be 16. 
%\delete{We choose identity mapping for the correlation embedding layer in attention fusion net. The attention mask prediction layer is implemented with a $3\times3$ conv layer followed by a non-linear function ReLU and a $1\times1$ conv layer.}
For the dynamic fusion net, we choose the same 4D conv layers as in the correlation network for the correlation embedding module, and the attention mask prediction layer is implemented with a $1\times1$ conv layer.

To train the model, we set $\lambda$ and $\gamma$ in the multi-auxiliary task loss to 1 by validation. The model parameters are randomly initialized except for feature extractor. The model is trained for 5 epochs on 4 GPUs with early stopping to avoid overfitting. We use Adam optimizer~\cite{kingma2014adam} with a learning rate of 5$\times$$10^{-4}$.

Images of all three datasets are first resized into the size of 400$\times$400. Our model is trained on the PF-PASCAL benchmark~\cite{ham2018proposal}. To further validate generalization capacity of our model, we test the trained model with the PF-WILLOW dataset~\cite{ham2016proposal} and the TSS dataset~\cite{ham2016proposal} without any further finetuning. Finally, we conduct the ablation study on the PF-PASCAL dataset~\cite{ham2018proposal}.

%Decide to not report TSS for now
%\subsection*{TSS(Taniai)\cite{taniai2016joint}}
%TSS dataset contains 400 image pairs in total, divided into three groups, including FG3DCAR,JODS and PASCAL. For FG3DCAr, there are 195 image pairs of automobiles. For JODS, there are 81 image pairs of airplanes, cars and horses. For PASCAL, there are 124 image pairs of trains, cars, buses, bikes, and motorbikes. Ground truth flows and foreground mask for image pair are provided, where we only use it for evaluation in weak supervision setting. On the condition $\alpha = 0.05$, we list previous PCK results to compare with the results of our method.

%\vspace{-2mm}

\subsection{PF-Pascal Benchmark}
\label{sec:pf-pascal-benchmark}

\paragraph{Dataset and Evaluation Metric} The PF-PASCAL~\cite{ham2018proposal} benchmark is built from the PASCAL 2011 keypoint annotation dataset~\cite{bourdev2009poselets}, which consists of 20 object categories. Following the dataset split in~\cite{khan2017}, we partition the total 1351 image pairs into a training set of 735 pairs, validation set of 308 pairs and test set of 308 pairs, respectively. The model learning is performed in a weakly-supervised manner where keypoint annotations are not used for training but for evaluation only. We report the percentage of the correct keypoints (PCK) metric~\cite{yang2013articulated} which measures the percentage of keypoints whose transfer errors below a given threshold. In line with previous work, we report PCK ($\alpha = 0.1$) w.r.t. image size.

\vspace{-3mm}
\paragraph{Experimental Results} %Table \ref{EPascal} shows the detailed comparison with previous methods for PCK $\alpha=0.1$. We can see that our method achieves the state-of-the-art results which outperforms the previous state-of-the-art method NC-Net \cite{Rocco18b} with $3.4\%$.

As shown in Table~\ref{EPascal}, we compare our proposed method with previous methods including NC-Net~\cite{Rocco18b}, WeakAlign~\cite{Rocco2018}, RTN~\cite{kim2018recurrent}, CNNGeo~\cite{Rocco2017}, Proposal Flow~\cite{ham2018proposal}, UCN~\cite{choy2016universal} and different versions of SCNet~\cite{khan2017}. Our approach achieves an overall PCK of $82.3\%$ , outperforming the prior state of the art~\cite{Rocco18b} by $3.4\%$.

\vspace{-3mm}
\paragraph{Visualization Results} Fig.~\ref{fig:vis_pfpascal} shows qualitative comparisons with Nc-Net~\cite{Rocco18b}. We can see that our model is robust against repetitive patterns thanks to our proposed context-aware semantic representation and dynamic fusion. More qualitative results can be found in the suppl. material.

%The leftmost column shows source images. The following columns shows predictions of model from our DCCNet,
 	
\begin{figure}[t]
	\centering
	\includegraphics[width=\linewidth]{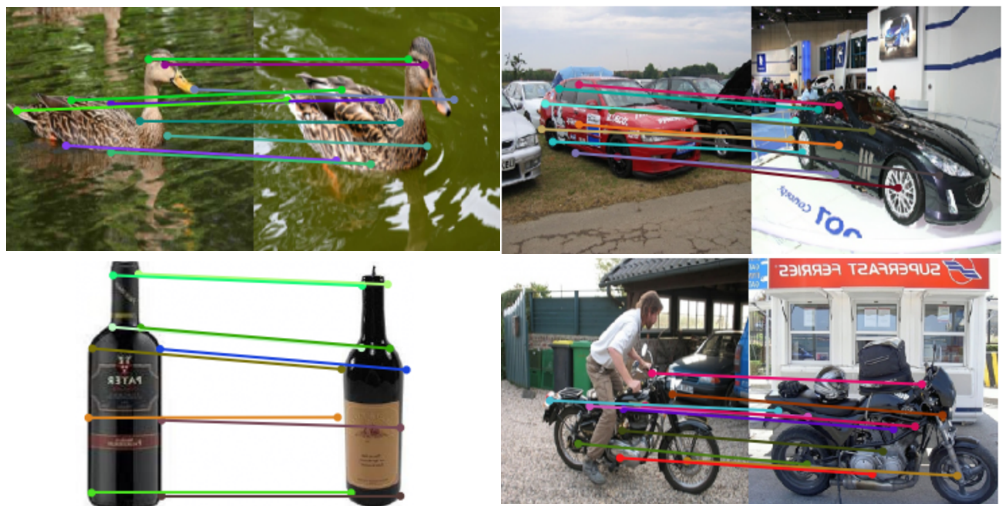}
	\caption{\textbf{Semantic alignment examples on PF-WILLOW.} Our model can produce reasonable matching results despite large background clutters and viewpoint changes.}\vspace{-3mm}
	\label{fig:vis_willow}
\end{figure}

\subsection{PF-WILLOW Benchmark}
\label{sec:pf-willow-benchmark}

\paragraph{Dataset and evaluation metric} The PF-WILLOW data-set consists of 900 image pairs selected from a total of 100 images~\cite{ham2016proposal}. 
%We report quantitative PCK $(\alpha = 0.05, 0.10, 0.15)$ results on this dataset.
We report the PCK scores with multiple thresholds $(\alpha = 0.05, 0.10, 0.15)$ w.r.t. bounding box size in order to compare with prior methods.

%4D_1layer,PCK0.05,0.10,0.15, WILLOW
\begin{table}[tb]
	\centering
	\resizebox{0.5\textwidth}{!}{
		\begin{tabular}{c|c c c}
			\hline
			%   & \multicolumn{3}{c}{Default} & \multicolumn{3}{c}{Know  Object}\\
			Methods  &$\alpha=0.05$&$\alpha=0.10$&$\alpha=0.15$\\  
			\hline
			\hline
			HOG+PF-LOM \cite{ham2018proposal} &28.4&56.8&68.2  \\
			DCTM \cite{kim2017dctm}  &38.1&61.0&72.1\\
			UCN-ST \cite{choy2016universal}  &24.1&54.0&66.5\\
			CAT-FCSS \cite{kim2018fcsscat} &36.2&54.6&69.2\\
			SCNet \cite{khan2017}  &38.6&70.4&85.3\\
			ResNet-101+CNNGeo \cite{Rocco2017}   &36.9&69.2&77.8\\
			ResNet-101+CNNGeo(W) \cite{Rocco2018}  &38.2&71.2&85.8\\
%			Yun-Chun \etal \cite{chendeep}  &[TODO]&[TODO]&[TODO]\\
			RTN \cite{kim2018recurrent}   &41.3&71.9&86.2\\
			NC-Net \cite{Rocco18b}   &\textbf{44.0}&72.7&85.4\\
			\hline
			Our Method           &43.6&\textbf{73.8}&\textbf{86.5}\\
			\hline
		\end{tabular}}
		\caption{\textbf{Evaluation results on PF-WILLOW~\cite{ham2016proposal}.} We report the PCK scores with three thresholds and the best results are in bold.} 
		\label{tab:evalwillow}
	\end{table}

\paragraph{Experimental Results} Table \ref{tab:evalwillow} compares the PCK accuracies of our DCCNet to those of the state-of-the-art semantic correspondence techniques. Our proposed method improves the PCK accuracies over the previously published best performance by $1.1\%$ when $\alpha = 0.10$ and $\alpha = 0.15$. Our model also achieves a competitive PCK ($\alpha = 0.05$) of $43.6\%$ which is merely $0.4\%$ lower than the state-of-the-art result, partially
due to the large scale variation in this dataset unseen in the training.  Fig.~\ref{fig:vis_willow} shows qualitative results on the PF-WILLOW dataset, which further demonstrate the strength of our method.
%which indicates that our attention module tends to reduce large errors.
%\delete{because that our representation is more suitable for coarse matching.}

\begin{figure}[tb]
 \centering
 \includegraphics[width=\linewidth]{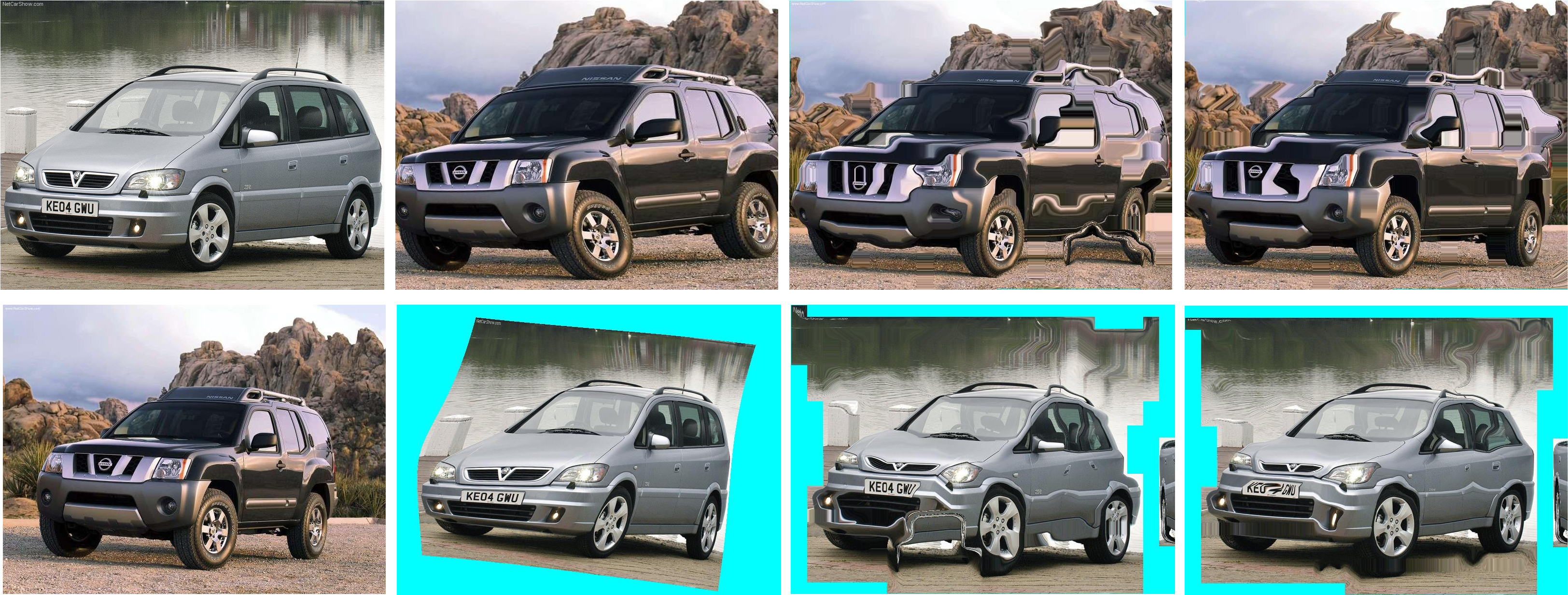}
% \caption{The first column of the images above is the input image pair from the TSS dataset. The second, third and fourth columns are respectively the output of WeakAlign, NC-Net and our method. The first row warps the image from source to target and the second row tries to warp the image from target to source. Compared to the other two methods, warped images from our model are smoother and more geometrically similar to the ground truth.}

 \caption{\textbf{Qualitative results on the TSS benchmark~\cite{taniai2016joint}.} The first column depicts source image and target image respectively. From the second column to the last column is results from WeakAlign \cite{Rocco2018}, NC-Net \cite{Rocco18b} and our model respectively.}
\vspace{-3mm}
 \label{fig:vis_tss}
\end{figure}

\vspace{-1mm}

\subsection{TSS Benchmark}
\label{sec:tss-benchmark}

\paragraph{Dataset and evaluation metric} The TSS dataset contains 400 image pairs in total, divided into three groups, including FG3DCAR, JODS, and PASCAL. Ground truth flows and foreground masks for image pair are provided, where we only use it for evaluation in the weak supervision setting. Following Taniai~\etal~\cite{taniai2016joint}, we report the PCK over foreground object by setting $\alpha$ to $0.05$ w.r.t. image size.

\vspace{-3mm}
\paragraph{Experimental Results} Table \ref{tabevalTSS} presents quantitative results on the TSS benchmark. We observe that our method outperforms previous methods on one of the three groups of the TSS dataset and our average performance over three groups on the TSS dataset achieves new state of the art. This shows our method can generalize to novel datasets despite the moderate change of data distribution. 
%\delete{It's expected that the gain of our method is modest because we apply our trained model on the dataset without finetuning and the training dataset itself is severely class imbalanced.} 
Qualitative results are presented in Fig.~\ref{fig:vis_tss}.
%This may be due to imbalanced classes on this dataset. Recall our model is trained on PF-PASCAL and directly test on TSS without any finetuning. 

\subsection{Ablation Study}
\label{sec:ablation-study}
%To demonstrate the effectiveness of the spatial context encoder, attention fusion net and multi-auxiliary task loss 
To understand the effectiveness of our model components, we conduct a series of ablation studies focusing on: 1) \textit{effects of individual modules}, 2) \textit{kernel sizes in spatial context}, 3) \textit{different fusion methods} and 4) \textit{multi-auxiliary task losses}. We select NC-Net~\cite{Rocco18b} as our baseline and report PCK ($\alpha=0.1$) on the PF-PASCAL \cite{ham2018proposal} test split.

\vspace{-3mm}
\paragraph{Effects of Individual Modules}
We consider five different ablation settings and the overall results are shown in Table~\ref{tab:ablation_sum}. First, we note that applying our proposed spatial context encoder (Baseline+S) generates large performance improvement ($2.0\%$) over NC-Net~\cite{Rocco18b}. Second, adding dynamic fusion with auxiliary loss (Baseline+SDA) provides a further boost of $1.4\%$. Below we introduce detailed analysis for each module via the other three ablation settings.

\begin{table}[tp]
	\centering
	\resizebox{0.45\textwidth}{!}{
		\begin{tabular}{c|c c c|c}
			\hline
			%   & \multicolumn{3}{c}{Default} & \multicolumn{3}{c}{Know  Object}\\
			Methods  &FG3D.&JODS&PASC.&avg.\\  
			\hline
			\hline
			HOG+PF-LOM \cite{ham2018proposal} &78.6&65.3&53.1&65.7 \\
			HOG+TSS \cite{taniai2016joint} &83.0&59.5&48.3&63.6  \\
			FCSS+SIFT Flow \cite{kim2017fcss}  &83.0&65.6&49.4&66.0\\
			FCSS+PF-LOM \cite{kim2017fcss}  &83.9&63.5&58.2&68.5\\
			HOG+OADSC \cite{yang2017object} &87.5&70.8&72.9&77.1\\
			FCSS+DCTM \cite{kim2017dctm}  &89.1&72.1&61.0&74.0\\
			VGG-16+CNNGeo \cite{Rocco2017}  &83.9&65.8&52.8&67.5\\
			ResNet-101+CNNGeo(S) \cite{Rocco2017}   &83.9&76.4&56.3&74.3\\
			ResNet-101+CNNGeo(W) \cite{Rocco2018}  &90.3&76.4&56.5&74.4\\
%		    Yun-Chun \etal \cite{chendeep}   &89.8&76.8&56.0&74.2\\
			RTN \cite{kim2018recurrent}   &90.1&78.2&\textbf{63.3}&77.2\\
			NC-Net \cite{Rocco18b}  &\textbf{94.5}&81.4&57.1&77.7\\
			\hline
			Our Method  &93.5&\textbf{82.6}&57.6&\textbf{77.9}\\
			\hline
	\end{tabular}}
	\caption{\textbf{Evaluation results on the TSS dataset~\cite{taniai2016joint}.} We report the PCK scores with $\alpha=0.05$ and the best results are in bold.}\vspace{-2mm}
	\label{tabevalTSS}
\end{table}

%\begin{table}[b]
%	\centering
%	\resizebox{0.25\textwidth}{!}{
%		\begin{tabular}{c|c}
%			\hline
%			Models  &Results  \\  
%			\hline
%			%\hline
%			NC-Net \cite{Rocco18b} &78.9 \\
%			Ours(+SCE) &80.9 \\
%			Ours(+SCE+DF)&\textbf{82.3} \\
%			\hline
%	\end{tabular}}
%	\caption{Effectiveness evaluation of our proposed components.}
%	\label{tab:components}
%\end{table}
\vspace{-3mm}
\paragraph{Spatial Context Encoder}
%\delete{We diagnose the influence of context-aware semantic representation kernel size. As shown in Table \ref{gskernel}, small kernel size shows limited effect. With increasing kernel size, matching performance significantly improves thanks to context-aware semantic representation which largely reduces matching ambiguities. However, further increasing kernel size leads to performance degrade due to induced noise. A potential strategy is to estimate a foreground mask to clean our context-aware semantic representation.}

Table~\ref{gskernel} shows the effects of incorporating context with different kernel sizes. For using our spatial context encoder alone (Baseline+S), the performance increases first and then drops with increasing kernel sizes, which is due to degradation of context-aware features as more background clutters are included. Our dynamic fusion and auxiliary loss (Baseline+SDA) can effectively alleviate the degradation problem.

\vspace{-3mm}
\paragraph{Fusion method}
We study the effects of our dynamic fusion by simple average fusion of two correlation maps, referring to the resulting model as Baseline+SAA. From Table~\ref{tab:ablation_sum} we can see that our dynamic fusion model (Baseline+SDA) yields significant better results ($82.3\%$) than average fusion ($80.2\%$), showing the necessity of our attention module. Moreover, Baseline+SAA underperforms the model setting with context-aware semantic feature alone (Baseline+S) due to its global averaging. In contrast, the pixel-wise weight mask from attention net enables each location to adaptively merge different scales of semantic cues. We also evaluate the model setting without correlation map embedding during dynamic fusion (Baseline+SCA), which generates worse results, indicating the efficacy of 4D correlation map features in the dynamic fusion network.

\begin{table}[tb]
	\centering
	\resizebox{0.49\textwidth}{!}{
		\begin{tabular}{l|c|c|c|c}
			\hline
			Models &SCE&Fusion&Auxiliary Loss&PCK  \\  
			\hline
			\hline
			NC-Net \cite{Rocco18b} &-&-&-&78.9 \\
			\hline
			Baseline+S&\cmark&-&-&80.9 \\
			Baseline+SCA&\cmark&Dynamic w/o Corr Embedding&\cmark&79.9 \\
			Baseline+SAA&\cmark&Average w/ Corr Embedding&\cmark&80.2 \\
			Baseline+SD&\cmark&Dynamic w/ Corr Embedding&\xmark&81.0 \\
			Baseline+SDA&\cmark&Dynamic w/ Corr Embedding&\cmark&\textbf{82.3} \\
			\hline
	\end{tabular}}
	\caption{\textbf{Analysis of individual modules of DCCNet on the PF-PASCAL \cite{ham2018proposal} dataset.} NC-Net~\cite{Rocco18b} is used as our baseline. Our ablation includes whether using spatial context encoder, fusion method adopted, and whether using multi-auxiliary task loss. }
	\label{tab:ablation_sum}
\end{table}

%  \begin{table}[ht]
%  	\centering
%  	\resizebox{0.22\textwidth}{!}{
%  		\begin{tabular}{c|c}
%  			\hline
%  			Fusion method  &Results  \\  
%  			\hline
%  			%\hline
%  			NC-Net \cite{Rocco18b} &78.9 \\
%  		    Ours(SCE)  &80.9 \\
%  			SCE+Average  &80.2 \\
%  			Ours(SCE+Attention) &\textbf{82.3} \\
%  			\hline
%  		\end{tabular}}
%  		\caption{Compare with average fusion method, necessity evaluation of dynamic fusion}
%  		\label{tab:fusion}
%  	\end{table}

\begin{table}[tb]
	\centering
	\resizebox{0.3\textwidth}{!}{
		\begin{tabular}{c c c}
			\hline
			%   & \multicolumn{3}{c}{Default} & \multicolumn{3}{c}{Know  Object}\\
			Models  &Kernel size &PCK  \\  
			\hline
			\hline
			NC-Net \cite{Rocco18b} &- &78.9 \\
			\hline
			Baseline+S  &11 &78.9 \\
			Baseline+S &25 &\textbf{80.9} \\
			Baseline+S &31 &77.1 \\
    {Baseline+SDA} &25 &\textbf{82.3} \\			
	{Baseline+SDA} &31 &80.7 \\
			\hline
		\end{tabular}}
% submission version:
%		\caption{Analysis of influence of different kernel size for proposed context-aware semantic representation. 
%		%+S indicates using proposed spatial context encoder. 
%		See text for detail.}

% final version draft:
		\caption{\textbf{Effect of kernel sizes in our spatial context on the PF-PASCAL \cite{ham2018proposal} dataset.} NC-Net~\cite{Rocco18b} is used as our baseline.}
		\label{gskernel}
	\end{table}
 \vspace{-2mm}

\paragraph{Multi-auxiliary task loss}
To validate the effect of our proposed auxiliary task loss, we train a model without two additional loss terms, which is referred to as Baseline+SD. Table~\ref{tab:ablation_sum} shows that our model with auxiliary loss terms (Baseline+SDA) attains $1.3\%$ higher PCK scores than the Baseline+SD model, reaching the state-of-the-art result of $82.3\%$. This improvement indicates the effectiveness of our multi-auxiliary task loss in regularizing the training process for weakly-supervised semantic correspondence task. With the multi-auxiliary task loss, our local feature and context-aware semantic feature branches have stronger supervision signals, which in turn benefits the fusion branch and produces better overall matching results.

%\begin{table}[tb]
%	\centering
%	\resizebox{0.5\textwidth}{!}{
%		\begin{tabular}{l|c|c|c|c}
%			\hline
%			Models &SCE&Fusion&Auxiliary Loss&PCK  \\  
%			\hline
%			\hline
%			NC-Net \cite{Rocco18b} &-&-&-&78.9 \\
%			\hline
%			Baseline+S&\cmark&-&-&80.9 \\
%			Baseline+SAA&\cmark&Average&\cmark&80.2 \\
%			Baseline+SD&\cmark&Dynamic&\xmark&81.0 \\
%			Baseline+SDA(DCCNet)&\cmark&Dynamic&\cmark&\textbf{82.3} \\
%			\hline
%		\end{tabular}}
%		\caption{\textbf{Analysis of each our proposed component on the PF-PASCAL \cite{ham2018proposal} dataset.} NC-Net~\cite{Rocco18b} is used as our baseline. From left to right is model, whether using spatial context encoder, fusion method adopted, whether using multi-auxiliary task loss and PCK results. }
%		\label{tab:ablation_sum}
%	\end{table}

\vspace{-1mm}

\section{Conclusion}
In this work, we have proposed an effective deep correspondence network, DCCNet, for the semantic alignment problem. Compared to the prior work, our approach has several innovations in semantic matching. First, we develop a learnable context-aware semantic representation that is robust against repetitive patterns and local ambiguities. In addition, we design a novel dynamic fusion module to adaptively combine semantic cues from multiple spatial scales. Finally, we adopt a multi-auxiliary task loss to better regularize the learning of our dynamic fusion strategy. We demonstrate the efficacy of our approach by extensive experimental evaluations on the PF-PASCAL, PF-WILLOW and TSS datasets. The results evidently show that our DCCNet achieves the superior or comparable performances over the prior state-of-the-art approaches on all three datasets.

\vspace{-2mm}
\paragraph{Acknowledgments}
This work was supported in part by the NSFC Grant No.61703195 and the Shanghai NSF Grant No.18ZR1425100.

{\small
\bibliographystyle{ieee_fullname}
\bibliography{weakalign.bib}

\begin{thebibliography}{10}\itemsep=-1pt

\bibitem{agarwal2011building}
Sameer Agarwal, Yasutaka Furukawa, Noah Snavely, Ian Simon, Brian Curless,
  Steven~M Seitz, and Richard Szeliski.
\newblock Building rome in a day.
\newblock {\em Communications of the ACM}, 54(10):105--112, 2011.

\bibitem{bourdev2009poselets}
Lubomir Bourdev and Jitendra Malik.
\newblock Poselets: Body part detectors trained using 3d human pose
  annotations.
\newblock In {\em Proceedings of the International Conference on Computer
  Vision(ICCV)}, 2009.

\bibitem{chen2017rethinking}
Liang-Chieh Chen, George Papandreou, Florian Schroff, and Hartwig Adam.
\newblock Rethinking atrous convolution for semantic image segmentation.
\newblock {\em arXiv preprint}, 2017.

\bibitem{chen2016attention}
Liang-Chieh Chen, Yi Yang, Jiang Wang, Wei Xu, and Alan~L Yuille.
\newblock Attention to scale: Scale-aware semantic image segmentation.
\newblock In {\em Proceedings of the IEEE Conference on Computer Vision and
  Pattern Recognition(CVPR)}, 2016.

\bibitem{chendeep}
Yun-Chun Chen, Po-Hsiang Huang, Li-Yu Yu, Jia-Bin Huang, Ming-Hsuan Yang, and
  Yen-Yu Lin.
\newblock Deep semantic matching with foreground detection and
  cycle-consistency.
\newblock In {\em Proceedings of the Asian Conference on Computer
  Vision(ACCV)}, 2018.

\bibitem{choy2016universal}
Christopher~B Choy, JunYoung Gwak, Silvio Savarese, and Manmohan Chandraker.
\newblock Universal correspondence network.
\newblock In {\em Advances in Neural Information Processing Systems(NeurIPS)},
  2016.

\bibitem{dale2009image}
Kevin Dale, Micah~K Johnson, Kalyan Sunkavalli, Wojciech Matusik, and Hanspeter
  Pfister.
\newblock Image restoration using online photo collections.
\newblock In {\em Proceedings of the International Conference on Computer
  Vision(ICCV)}, 2009.

\bibitem{ham2016proposal}
Bumsub Ham, Minsu Cho, Cordelia Schmid, and Jean Ponce.
\newblock Proposal flow.
\newblock In {\em Proceedings of the IEEE Conference on Computer Vision and
  Pattern Recognition(CVPR)}, 2016.

\bibitem{ham2018proposal}
Bumsub Ham, Minsu Cho, Cordelia Schmid, and Jean Ponce.
\newblock Proposal flow: Semantic correspondences from object proposals.
\newblock {\em IEEE Transactions on Pattern Analysis and Machine Intelligence},
  2018.

\bibitem{han2017scnet}
Kai Han, Rafael~S Rezende, Bumsub Ham, Kwan-Yee~K Wong, Minsu Cho, Cordelia
  Schmid, and Jean Ponce.
\newblock Scnet: Learning semantic correspondence.
\newblock In {\em Proceedings of the IEEE Conference on Computer Vision and
  Pattern Recognition(CVPR)}, 2017.

\bibitem{khan2017}
Kai Han, Rafael~S. Rezende, Bumsub Ham, Kwan-Yee~K. Wong, Minsu Cho, Cordelia
  Schmid, and Jean Ponce.
\newblock Scnet: Learning semantic correspondence.
\newblock In {\em Proceedings of the International Conference on Computer
  Vision(ICCV)}, 2017.

\bibitem{he2016deep}
Kaiming He, Xiangyu Zhang, Shaoqing Ren, and Jian Sun.
\newblock Deep residual learning for image recognition.
\newblock In {\em Proceedings of the IEEE Conference on Computer Vision and
  Pattern Recognition(CVPR)}, 2016.

\bibitem{hirschmuller2007stereo}
Heiko Hirschmuller.
\newblock Stereo processing by semiglobal matching and mutual information.
\newblock {\em IEEE Transactions on pattern analysis and machine intelligence},
  30(2):328--341, 2007.

\bibitem{hongsuck2018attentive}
Paul Hongsuck~Seo, Jongmin Lee, Deunsol Jung, Bohyung Han, and Minsu Cho.
\newblock Attentive semantic alignment with offset-aware correlation kernels.
\newblock In {\em Proceedings of the European Conference on Computer
  Vision(ECCV)}, 2018.

\bibitem{horn1981determining}
Berthold~KP Horn and Brian~G Schunck.
\newblock Determining optical flow.
\newblock {\em Artificial intelligence}, 17(1-3):185--203, 1981.

\bibitem{jeon2018parn}
Sangryul Jeon, Seungryong Kim, Dongbo Min, and Kwanghoon Sohn.
\newblock Parn: Pyramidal affine regression networks for dense semantic
  correspondence.
\newblock In {\em Proceedings of the European Conference on Computer
  Vision(ECCV)}, 2018.

\bibitem{kim2019laf}
Sunok Kim, Seungryong Kim, Dongbo Min, and Kwanghoon Sohn.
\newblock Laf-net: Locally adaptive fusion networks for stereo confidence
  estimation.
\newblock In {\em Proceedings of the IEEE Conference on Computer Vision and
  Pattern Recognition(CVPR)}, 2019.

\bibitem{kim2018recurrent}
Seungryong Kim, Stephen Lin, SANG~RYUL JEON, Dongbo Min, and Kwanghoon Sohn.
\newblock Recurrent transformer networks for semantic correspondence.
\newblock In {\em Advances in Neural Information Processing Systems(NeurIPS)},
  2018.

\bibitem{kim2017fcss}
Seungryong Kim, Dongbo Min, Bumsub Ham, Sangryul Jeon, Stephen Lin, and
  Kwanghoon Sohn.
\newblock Fcss: Fully convolutional self-similarity for dense semantic
  correspondence.
\newblock In {\em Proceedings of the IEEE Conference on Computer Vision and
  Pattern Recognition(CVPR)}, 2017.

\bibitem{kim2018fcsscat}
Seungryong Kim, Dongbo Min, Bumsub Ham, Stephen Lin, and Kwanghoon Sohn.
\newblock Fcss: Fully convolutional self-similarity for dense semantic
  correspondence.
\newblock In {\em IEEE Transactions on Pattern Analysis and Machine
  Intelligence}, 2018.

\bibitem{kim2015dasc}
Seungryong Kim, Dongbo Min, Bumsub Ham, Seungchul Ryu, Minh~N Do, and Kwanghoon
  Sohn.
\newblock Dasc: Dense adaptive self-correlation descriptor for multi-modal and
  multi-spectral correspondence.
\newblock In {\em Proceedings of the IEEE Conference on Computer Vision and
  Pattern Recognition(CVPR)}, 2015.

\bibitem{kim2016deep}
Seungryong Kim, Dongbo Min, Stephen Lin, and Kwanghoon Sohn.
\newblock Deep self-correlation descriptor for dense cross-modal
  correspondence.
\newblock In {\em Proceedings of the European Conference on Computer
  Vision(ECCV)}, 2016.

\bibitem{kim2017dctm}
Seungryong Kim, Dongbo Min, Stephen Lin, and Kwanghoon Sohn.
\newblock Dctm: Discrete-continuous transformation matching for semantic flow.
\newblock In {\em Proceedings of the IEEE Conference on Computer Vision and
  Pattern Recognition(CVPR)}, 2017.

\bibitem{kingma2014adam}
Diederik~P Kingma and Jimmy Ba.
\newblock Adam: A method for stochastic optimization.
\newblock In {\em Proceedings of the International Conference on Learning
  Representations(ICLR)}, 2014.

\bibitem{liu2011sift}
Ce Liu, Jenny Yuen, and Antonio Torralba.
\newblock Sift flow: Dense correspondence across scenes and its applications.
\newblock {\em IEEE Transactions on Pattern Analysis and Machine Intelligence},
  2011.

\bibitem{luo2016understanding}
Wenjie Luo, Yujia Li, Raquel Urtasun, and Richard Zemel.
\newblock Understanding the effective receptive field in deep convolutional
  neural networks.
\newblock In {\em Advances in Neural Information Processing Systems(NeurIPS)},
  2016.

\bibitem{nikandrova2015category}
Ekaterina Nikandrova and Ville Kyrki.
\newblock Category-based task specific grasping.
\newblock {\em Robotics and Autonomous Systems}, 2015.

\bibitem{novotny2017anchornet}
David Novotny, Diane Larlus, and Andrea Vedaldi.
\newblock Anchornet: A weakly supervised network to learn geometry-sensitive
  features for semantic matching.
\newblock In {\em Proceedings of the IEEE Conference on Computer Vision and
  Pattern Recognition(CVPR)}, 2017.

\bibitem{paszke2017automatic}
Adam Paszke, Sam Gross, Soumith Chintala, Gregory Chanan, Edward Yang, Zachary
  DeVito, Zeming Lin, Alban Desmaison, Luca Antiga, and Adam Lerer.
\newblock Automatic differentiation in pytorch.
\newblock 2017.

\bibitem{Rocco2017}
Ignacio Rocco, Relja Arandjelovi{\'c}, and Josef Sivic.
\newblock Convolutional neural network architecture for geometric matching.
\newblock In {\em Proceedings of the IEEE Conference on Computer Vision and
  Pattern Recognition(CVPR)}, 2017.

\bibitem{Rocco2018}
Ignacio Rocco, Relja Arandjelovi{\'c}, and Josef Sivic.
\newblock End-to-end weakly-supervised semantic alignment.
\newblock In {\em Proceedings of the IEEE Conference on Computer Vision and
  Pattern Recognition(CVPR)}, 2018.

\bibitem{Rocco18b}
Ignacio Rocco, Mircea Cimpoi, Relja Arandjelovi\'{c}, Akihiko Torii, Tomas
  Pajdla, and Josef Sivic.
\newblock Neighbourhood consensus networks.
\newblock In {\em Advances in Neural Information Processing Systems(NeurIPS)},
  2018.

\bibitem{scharstein2002taxonomy}
Daniel Scharstein and Richard Szeliski.
\newblock A taxonomy and evaluation of dense two-frame stereo correspondence
  algorithms.
\newblock {\em International journal of computer vision}, 47(1-3):7--42, 2002.

\bibitem{shechtman2007matching}
Eli Shechtman and Michal Irani.
\newblock Matching local self-similarities across images and videos.
\newblock In {\em Proceedings of the IEEE Conference on Computer Vision and
  Pattern Recognition(CVPR)}, 2007.

\bibitem{taniai2016joint}
Tatsunori Taniai, Sudipta~N Sinha, and Yoichi Sato.
\newblock Joint recovery of dense correspondence and cosegmentation in two
  images.
\newblock In {\em Proceedings of the IEEE Conference on Computer Vision and
  Pattern Recognition(CVPR)}, 2016.

\bibitem{tola2010daisy}
Engin Tola, Vincent Lepetit, and Pascal Fua.
\newblock Daisy: An efficient dense descriptor applied to wide-baseline stereo.
\newblock {\em IEEE Transactions on Pattern Analysis and Machine Intelligence},
  2010.

\bibitem{wang2018non}
Xiaolong Wang, Ross Girshick, Abhinav Gupta, and Kaiming He.
\newblock Non-local neural networks.
\newblock In {\em Proceedings of the IEEE Conference on Computer Vision and
  Pattern Recognition}, pages 7794--7803, 2018.

\bibitem{yang2017object}
Fan Yang, Xin Li, Hong Cheng, Jianping Li, and Leiting Chen.
\newblock Object-aware dense semantic correspondence.
\newblock In {\em Proceedings of the IEEE Conference on Computer Vision and
  Pattern Recognition(CVPR)}, 2017.

\bibitem{yang2013articulated}
Yi Yang and Deva Ramanan.
\newblock Articulated human detection with flexible mixtures of parts.
\newblock {\em IEEE Transactions on Pattern Analysis and Machine Intelligence},
  2013.

\end{thebibliography}
}

\end{document}